\documentclass[10pt,a4paper]{article}
\usepackage[margin=1in]{geometry}
\usepackage[T1]{fontenc}
\usepackage[utf8]{inputenc}
\usepackage{lmodern}
\usepackage{microtype}
\usepackage{amsmath,amssymb,mathtools}
\usepackage{booktabs,tabularx,longtable,array,multirow}
\usepackage{enumitem}
\usepackage{xcolor}
\usepackage{graphicx}
\usepackage{tikz}
\usetikzlibrary{arrows.meta,positioning,fit,calc,shapes.geometric,backgrounds}
\usepackage[most]{tcolorbox}
\usepackage{caption}
\usepackage{subcaption}
\usepackage{float}
\usepackage{titlesec}
\usepackage{natbib}
\usepackage[hidelinks]{hyperref}
\usepackage[nameinlink,noabbrev]{cleveref}
\usepackage{fancyhdr}
\usepackage{lastpage}

\definecolor{blueA}{HTML}{1F4E79}
\definecolor{blueB}{HTML}{D9EAF7}
\definecolor{greenA}{HTML}{2E6B4F}
\definecolor{greenB}{HTML}{E3F2E9}
\definecolor{orangeA}{HTML}{A85B00}
\definecolor{orangeB}{HTML}{FCEBD5}
\definecolor{purpleA}{HTML}{6A4C93}
\definecolor{purpleB}{HTML}{EEE7F7}
\definecolor{grayA}{HTML}{4A4A4A}
\definecolor{grayB}{HTML}{F2F2F2}
\definecolor{redA}{HTML}{9B2C2C}
\definecolor{redB}{HTML}{F9E3E3}

\hypersetup{colorlinks=true,linkcolor=blueA,citecolor=greenA,urlcolor=blueA}
\setlist[itemize]{leftmargin=*,topsep=2pt,itemsep=1pt,parsep=0pt}
\setlist[enumerate]{leftmargin=*,topsep=2pt,itemsep=1pt,parsep=0pt}

\titleformat{\section}{\large\bfseries\color{blueA}}{\thesection}{0.55em}{}
\titleformat{\subsection}{\normalsize\bfseries\color{grayA}}{\thesubsection}{0.55em}{}
\titleformat{\subsubsection}{\normalsize\bfseries}{\thesubsubsection}{0.55em}{}
\newcommand{\R}{\mathbb{R}}
\newcommand{\SO}{\mathrm{SO}}
\newcommand{\OO}{\mathrm{O}}

\newcommand{\E}{\mathbb{E}}
\newcommand{\vect}[1]{\boldsymbol{#1}}
\newcommand{\doi}[1]{\href{https://doi.org/#1}{doi:#1}}
\newcolumntype{P}[1]{>{\raggedright\arraybackslash}p{#1}}
\newtcolorbox{keybox}[1]{
  colback=blueB,colframe=blueA,fonttitle=\bfseries,title=#1,
  boxrule=0.6pt,arc=2pt,left=6pt,right=6pt,top=5pt,bottom=5pt
}
\newtcolorbox{cautionbox}[1]{
  colback=orangeB,colframe=orangeA,fonttitle=\bfseries,title=#1,
  boxrule=0.6pt,arc=2pt,left=6pt,right=6pt,top=5pt,bottom=5pt
}
\newtcolorbox{gapbox}[1]{
  colback=purpleB,colframe=purpleA,fonttitle=\bfseries,title=#1,
  boxrule=0.7pt,arc=2pt,left=6pt,right=6pt,top=5pt,bottom=5pt
}
\newtcolorbox{successbox}[1]{
  colback=greenB,colframe=greenA,fonttitle=\bfseries,title=#1,
  boxrule=0.7pt,arc=2pt,left=6pt,right=6pt,top=5pt,bottom=5pt
}

\title{\vspace{-1.2cm}\textbf{Physically Typed and Geometry-Aware Representations for Earth Foundation Models}\\
\large A falsifiable research program for combining semantic geoembeddings with scalar, vector, axial/parity-sensitive, and tensor fields}
\author{Rajiv Ranjan\\Plaksha University\\\texttt{rajiv.ranjan@plaksha.edu.in}}
\date{}

\begin{document}
\maketitle
\vspace{-0.5cm}

\begin{abstract}
Earth-observation (EO) foundation models have become exceptionally effective at learning semantic, high-dimensional geospatial embeddings, while modern weather and climate models have demonstrated that Earth-specific geometry, spherical operators, meshes, and hybrid physical solvers can materially improve prediction. Yet these two advances are not equivalent. A conventional latent embedding has no inherent physical transformation law, whereas scalar fields, tangent polar-vector fields, axial/pseudovector quantities, covectors, and higher-order tensors transform differently under rotations, reflections, and changes of local coordinate frame. This proposal asks whether a general-purpose Earth foundation model should preserve those distinctions explicitly, or whether standard embeddings plus augmentation already learn everything that matters. The central contribution is therefore not a more complicated architecture by assumption, but a staged falsification program. A compute-conscious ERA5 dry run first compares conventional, augmentation-matched, typed-equivariant, and Hodge/Helmholtz variants under spatial, temporal, orientation, and low-data shifts. Only if explicit geometric typing yields reproducible improvements does the program advance toward a multimodal Earth foundation model in which semantic embeddings coexist with physically typed fields. The proposed gap is narrower and more defensible than claiming that current models ignore geometry entirely: several systems already respect spherical domain geometry, and emerging work explicitly learns scalar/vector fields on spheres. The unresolved question is whether \emph{foundation-scale, multimodal, parity-aware field typing} produces practical gains beyond those existing approaches.
\end{abstract}

\begin{keybox}{Executive conclusion}
The research gap is \textbf{realistic and scientifically important, but conditional}. The strongest defensible claim is not ``existing Earth FMs use only embeddings,'' because weather models already exploit spherical geometry, graph meshes, physical solvers, and vector-valued atmospheric channels. The gap is that current general-purpose EO/Earth FMs do not appear to maintain a systematic, explicit representation algebra in which scalar, tangent-vector, parity/axial, and higher-order tensor fields keep their correct transformation laws while also participating in a transferable semantic latent space. Recent scalar/vector-on-sphere work shows feasibility but is not yet a multimodal foundation-model solution \citep{ballerin,delouis}. Therefore the thesis should begin with the smallest experiment capable of disproving the value of explicit field typing.
\end{keybox}

\section{Conceptual foundation: not all ``vectors'' are the same}

The word \emph{vector} is overloaded in machine learning. A 768-dimensional transformer embedding is a vector in a linear-algebra sense, but it is not automatically a physical vector. A physical wind vector is tied to geographic space and must rotate when the coordinate frame rotates. A learned embedding has no such transformation semantics unless the architecture explicitly imposes them. This distinction is the foundation of the proposed research.

\begin{table}[H]
\centering
\caption{Geometric object types relevant to GeoAI and Earth-system modeling. Here $R$ denotes a rotation or orthogonal transformation and $\det(R)$ distinguishes proper rotations from reflections.}
\label{tab:types}
\small
\begin{tabularx}{\textwidth}{P{2.4cm} P{3.2cm} P{3.6cm} >{\raggedright\arraybackslash}X}
\toprule
\textbf{Object} & \textbf{Transformation idea} & \textbf{Earth examples} & \textbf{Why it matters to ML} \\
\midrule
Scalar / 0th-order tensor & $s' = s$ under frame rotation & temperature, pressure, elevation, reflectance, NDVI & Can be safely represented as ordinary scalar channels when coordinate frame changes. \\
Polar / true vector & $\vect v' = R\vect v$ & wind, ocean current, horizontal displacement, optical flow & Components must rotate together. Treating $(u,v)$ as unrelated channels can break frame consistency. \\
Axial vector / pseudovector & $\vect a' = \det(R)R\vect a$ & angular velocity, 3D vorticity, magnetic field & Indistinguishable from polar vectors under proper rotations, but differs under reflections/parity. \\
2D pseudoscalar & $\zeta'=\det(R)\zeta$ & vertical component of atmospheric/oceanic vorticity on a local tangent plane & Important correction: on a 2D Earth-surface patch, signed vertical vorticity is more naturally a pseudoscalar than a full 3D axial vector. \\
Covector / 1-form & transforms with inverse-transpose action & gradients, differentials, flux-related dual quantities & Useful when distinguishing ``direction of motion'' from ``linear functional on directions''; relevant in differential-geometry formulations. \\
Rank-2 tensor & $T'=RTR^{\top}$ (for a common tensor action) & stress/strain, diffusion tensor, structure tensor, SAR coherency/covariance objects & Encodes anisotropy and directional interactions that cannot be reduced to a single vector without loss. \\
Differential $k$-form & pullback under coordinate changes & circulation/flux-oriented discretizations, conservation laws & Natural language for discrete exterior calculus and conservation-aware learning. \\
Vector/tensor \emph{field} & geometric object attached to every spatial point & global wind field, currents, strain field & Requires local frames on curved surfaces and consistent transport/comparison between locations. \\
Learned embedding & usually $\vect z\in\R^d$ with no prescribed physical group action & AlphaEarth, TESSERA, Clay, and Prithvi feature vectors & Excellent semantic representation; physical interpretation of individual dimensions is usually absent. It \emph{can} be made typed/equivariant, but conventional embeddings are not. \\
\bottomrule
\end{tabularx}
\end{table}

\begin{figure}[H]
\centering
\begin{tikzpicture}[node distance=7mm and 8mm,>=Latex,every node/.style={font=\small}]
\node[draw,rounded corners,fill=grayB,minimum width=3.1cm,minimum height=1.05cm] (obs) {Earth observations};
\node[draw,rounded corners,fill=greenB,right=of obs,minimum width=3.3cm,minimum height=1.05cm] (typed) {Typed physical fields};
\node[draw,rounded corners,fill=blueB,right=of typed,minimum width=3.4cm,minimum height=1.05cm] (latent) {Learned semantic latent};
\node[draw,rounded corners,fill=purpleB,right=of latent,minimum width=3.1cm,minimum height=1.05cm] (tasks) {Downstream tasks};
\draw[->,thick] (obs) -- node[above]{interpret} (typed);
\draw[->,thick] (typed) -- node[above]{encode / interact} (latent);
\draw[->,thick] (latent) -- node[above]{adapt} (tasks);
\node[below=5mm of typed,align=center,text width=3.3cm] {scalar $\oplus$ vector $\oplus$\\ pseudoscalar/axial $\oplus$ tensor};
\node[below=5mm of latent,align=center,text width=3.4cm] {generic invariant tokens\\ + typed equivariant channels};
\node[below=5mm of obs,align=center,text width=3.1cm] {S1/S2, ERA5, DEM,\\ ocean, text, labels};
\node[below=5mm of tasks,align=center,text width=3.1cm] {forecasting, mapping,\\ retrieval, change, extremes};
\end{tikzpicture}
\caption{Proposed representation ladder. The core hypothesis is not that semantic embeddings should be replaced, but that physically typed fields should coexist with embeddings where transformation laws carry useful information.}
\label{fig:ladder}
\end{figure}
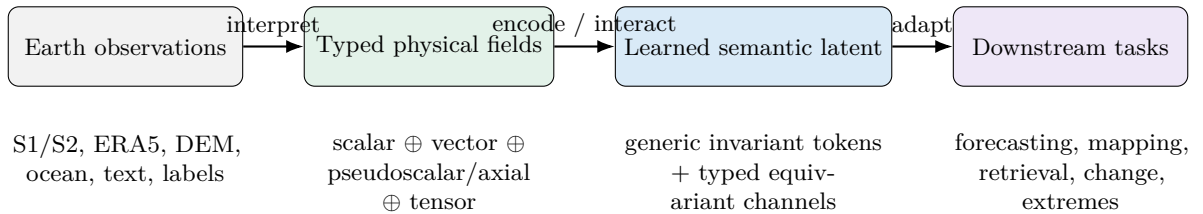

\subsection{Dot products, cross products, and why this matters for learned representations}
For embeddings $\vect z_1,\vect z_2$, cosine similarity is a normalized dot product,
\begin{equation}
\mathrm{cosim}(\vect z_1,\vect z_2)=\frac{\vect z_1\cdot\vect z_2}{\|\vect z_1\|\,\|\vect z_2\|},
\end{equation}
whose semantic meaning is learned through the training objective. For two 3D polar vectors, the cross product $\vect a\times\vect b$ is axial; its normalized magnitude is $|\sin\theta|$, a measure of perpendicularity rather than similarity. In dimensions other than three, the cross product is not generally a vector; exterior/wedge products are the appropriate generalization. This is one reason a generic high-dimensional embedding should not be casually interpreted using physical-vector language.

\section{Evolving literature: from semantic embeddings to geometric Earth models}

\subsection{Phase I: EO foundation models optimize semantic transfer}
Masked autoencoding and multimodal self-supervision established the modern EO foundation-model paradigm. SatMAE exploits temporal and multispectral structure \citep{satmae}; Prithvi and Prithvi-EO-2.0 pretrain multi-temporal HLS transformers and add temporal/location information \citep{prithvi,prithvi2}; CROMA aligns radar and optical imagery \citep{croma}; DOFA adapts a shared model across wavelengths and sensors \citep{dofa}; Galileo jointly represents optical, SAR, elevation, weather, pseudo-labels, and other modalities across space and time \citep{galileo}; TerraMind learns any-to-any multimodal generation across nine geospatial modalities \citep{terramind}; and Clay explicitly exposes semantic geospatial embeddings from imagery plus location/time metadata \citep{clay}.

The most recent embedding-field systems push this idea further. AlphaEarth Foundations (AEF) produces compact annual global embedding fields and explicitly describes its outputs as 64-byte embeddings constrained on $S^{63}$; its inputs are encoded source frames transformed into a common latent space \citep{alphaearth}. TESSERA v2 studies the scaling of pixel-wise EO embeddings and reports Matryoshka representations that can be truncated for storage/accuracy trade-offs \citep{tessera2}. OlmoEarth continues the trend toward stable multimodal spatiotemporal latent-image modeling \citep{olmoearth}.

These models establish a powerful fact: \textbf{generic latent semantics transfer extremely well}. They do not, however, establish that all physically different measurements should share the same transformation behavior once inside the model.

\begingroup\setlength{\tabcolsep}{4pt}\small
\begin{longtable}{P{1.8cm} P{2.35cm} P{2.45cm} P{2.35cm} P{1.8cm} P{3.25cm}}
\caption{Representative EO foundation models and the distinction between semantic representation and explicit physical field typing. ``Not documented'' means the cited architecture does not claim polar/axial/tensor transformation semantics; it does not mean the model cannot learn useful directional patterns.}\label{tab:eofm}\\
\toprule
\textbf{Model} & \textbf{Primary inputs} & \textbf{Representation emphasis} & \textbf{Geometry / context} & \textbf{Explicit typed fields?} & \textbf{Interpretation for this proposal} \\
\midrule
\endfirsthead
\toprule
\textbf{Model} & \textbf{Primary inputs} & \textbf{Representation emphasis} & \textbf{Geometry / context} & \textbf{Explicit typed fields?} & \textbf{Interpretation for this proposal} \\
\midrule
\endhead
SatMAE \citep{satmae} & temporal + multispectral satellite imagery & MAE latent tokens & temporal and spectral positional encodings & Not documented & Strong EO-specific SSL, but physical object type is not central. \\
Prithvi-EO-2.0 \citep{prithvi2} & global HLS time series & transformer embeddings & temporal and location embeddings & Not documented & Geographic metadata improves representation without requiring field-type algebra. \\
CROMA \citep{croma} & Sentinel-1 + Sentinel-2 & radar/optical unimodal and fused embeddings & cross-modal contrastive + 2D spatial bias & Not documented & Multimodality is handled semantically. \\
DOFA \citep{dofa} & multisensor spectral imagery & shared transformer features & wavelength-conditioned hypernetwork & Not documented & Sensor physics enters through wavelengths, not vector/tensor transformation laws. \\
Clay \citep{clay} & imagery + location + time (+ sensor metadata) & semantic embeddings & geospatial + temporal metadata in ViT/MAE framework & Not documented & Explicitly an embedding-producing Earth model. \\
Galileo \citep{galileo} & optical, SAR, elevation, weather, pseudo-labels, more & global + local self-supervised features & multi-scale spatiotemporal masking & Not documented & Particularly relevant because weather is included, yet modalities are unified through representation learning rather than typed field irreps. \\
TerraMind \citep{terramind} & nine EO modalities & token- and pixel-scale generative representations & dual-scale multimodal fusion & Not documented & Strong multimodal generative FM; provides a natural future host for typed adapters. \\
AlphaEarth \citep{alphaearth} & multiple EO and environmental sources + time & compact annual embedding field & continuous time conditioning, sensor metadata, spatially precise latent bottleneck & Not documented & Closest exemplar of planet-scale semantic ``field'' embeddings; ``embedding field'' is not the same as a physical vector field. \\
TESSERA v2 \citep{tessera2} & Sentinel-1/2 annual time series & 128-D pixel embeddings / Matryoshka prefixes & temporal SSL and scaling/distillation & Not documented & Demonstrates that compact embeddings can be extremely strong; a geometry-aware model must beat this simplicity where it claims benefit. \\
OlmoEarth \citep{olmoearth} & multimodal spatiotemporal EO & latent image representation & EO-specific masking/loss design & Not documented & Strong 2025 multimodal baseline for future representation comparisons. \\
\bottomrule
\end{longtable}
\endgroup

\subsection{Phase II: weather and climate models make Earth geometry impossible to ignore}
Weather models expose a different side of the problem because the state contains true directional quantities. FourCastNet forecasts global atmospheric variables with Fourier neural operators \citep{fourcastnet}; Pangu-Weather incorporates a 3D Earth-specific transformer and vertical structure \citep{pangu}; GraphCast maps between latitude--longitude grids and an icosahedral multimesh using a GNN \citep{graphcast}; GenCast adds probabilistic generative forecasting \citep{gencast}; Aurora pretrains a 1.3-billion-parameter Earth-system model over heterogeneous geophysical datasets \citep{aurora}; and NeuralGCM combines a differentiable dynamical core with learned physics \citep{neuralgcm}.

This literature already invalidates a simplistic claim that Earth AI ignores geometry. In particular, SFNO explicitly constructs an $\SO(3)$-equivariant spherical operator for functions on $S^2$ \citep{sfno}. However, there is an important second distinction: \textbf{domain equivariance is not automatically field-type equivariance}. SFNO formulates atmospheric state as a vector of channels $u(x,t)\in\R^N$ over the sphere and mixes those channels; the standard pullback action used for scalar functions does not by itself encode that horizontal wind components form a tangent vector whose basis changes over the sphere. Likewise, Aurora states that input variables are treated as $H\times W$ images and mapped with variable-specific linear transformations before entering a universal latent representation \citep{aurora}.

\begin{table}[H]
\centering
\caption{Earth-system models: increasingly geometry/physics-aware, but not generally explicit about heterogeneous field transformation types.}
\label{tab:earthmodels}
\scriptsize
\begin{tabularx}{\textwidth}{p{1.9cm} p{2.7cm} p{2.3cm} p{2.6cm} p{2.2cm} X}
\toprule
\textbf{Model} & \textbf{State / variables} & \textbf{Domain geometry} & \textbf{Internal representation} & \textbf{Explicit field-type handling} & \textbf{Key implication} \\
\midrule
FourCastNet \citep{fourcastnet} & multichannel ERA5 incl. wind & lat--lon / Fourier & AFNO latent channels & No documented polar/axial irrep typing & Fast global operator; flat Fourier geometry motivated later spherical variants. \\
Pangu-Weather \citep{pangu} & 69 factors incl. upper-air and surface variables & 3D Earth-specific priors & 3D transformer features & No documented field-type irrep typing & Strong evidence that Earth-specific structural priors matter. \\
GraphCast \citep{graphcast} & hundreds of weather targets including vector-component variables & grid + icosahedral multimesh & node/edge feature vectors in GNN & Geometry-aware graph; no documented polar/axial representation algebra & Graph ``typing'' in the code refers to node/edge graph types, not necessarily physical tensor-field irreps. \\
GenCast \citep{gencast} & probabilistic global weather state & global graph/spherical weather structure & diffusion/generative latent state & No documented field-type irrep typing & Probabilistic uncertainty can be combined with, but is orthogonal to, geometric field typing. \\
Aurora \citep{aurora} & heterogeneous atmospheric, wave, chemistry variables & latitude--longitude + vertical latent levels & Perceiver encoder $\rightarrow$ 3D Swin latent $\rightarrow$ decoder & Variables tokenized as images; no documented polar/axial typing & A highly relevant FM baseline: heterogeneous variables are unified by embeddings. \\
SFNO / FCNv2 \citep{sfno} & $N$ physical channels over $S^2$ & explicit sphere; $\SO(3)$-equivariant convolution in continuous limit & spherical harmonic latent operator & Sphere equivariance for channel-valued functions, but not explicit tangent-vector/parity typing & Critical bridge: proves geometry can improve stability while leaving open richer field representations. \\
NeuralGCM \citep{neuralgcm} & atmospheric state in a dynamical solver & spherical spectral dynamical core & physical state + learned parameterization & Governing-equation variables handled physically in core, not a generic typed FM latent & Shows hybrid physics can outperform purely generic representation design for long-term dynamics. \\
\bottomrule
\end{tabularx}
\end{table}

\subsection{Phase III: geometric deep learning supplies the missing representation algebra}
Tensor Field Networks demonstrate layers whose inputs/outputs can be scalars, vectors, and higher-order tensors while preserving equivariance \citep{tfn}; SE(3)-Transformers extend equivariance to attention \citep{se3}; and the broader geometric deep-learning framework formalizes symmetry as an architectural prior \citep{gdl}. O(3)-equivariant molecular models such as NequIP explicitly use irreducible representations carrying both rotation order and parity, with large gains in data efficiency in their domain \citep{nequip}. Gauge-equivariant CNNs show how local frames can be handled on manifolds and specifically demonstrate climate-pattern applications on an icosahedral approximation to the sphere \citep{gauge}.

Most importantly for novelty assessment, recent work directly addresses scalar and vector fields on spheres. Ballerin et al. propose $\SO(3)$-equivariant networks for scalar/vector spherical signals such as temperature and wind \citep{ballerin}. A 2026 EGU contribution evaluates gauge-equivariant HEALPix U-Nets for global SST, emphasizing local orientation \citep{delouis}. These papers reduce the novelty of merely ``using equivariance on Earth.'' They strengthen a more precise gap: scaling typed geometric fields into a \emph{general-purpose multimodal foundation representation} and testing whether the added structure remains useful after strong augmentation, large-scale pretraining, and semantic fusion.

\subsection{Phase IV: negative evidence argues for a falsifiable rather than celebratory proposal}
Two results are especially important for research discipline. First, TESSERA v2 reports that pretraining loss is only weakly predictive of downstream performance and that carefully scaled/distilled compact embeddings can outperform much larger alternatives \citep{tessera2}. Second, SwissCrop25 reports an operational multi-year crop-mapping setting in which domain-specific time-series models outperform Galileo \citep{swisscrop}. Thus ``foundation model'' or ``more structure'' cannot be assumed to dominate task-specific simplicity. A successful geometry-aware FM must demonstrate gains under conditions where the geometry is causally relevant, not merely add parameters.

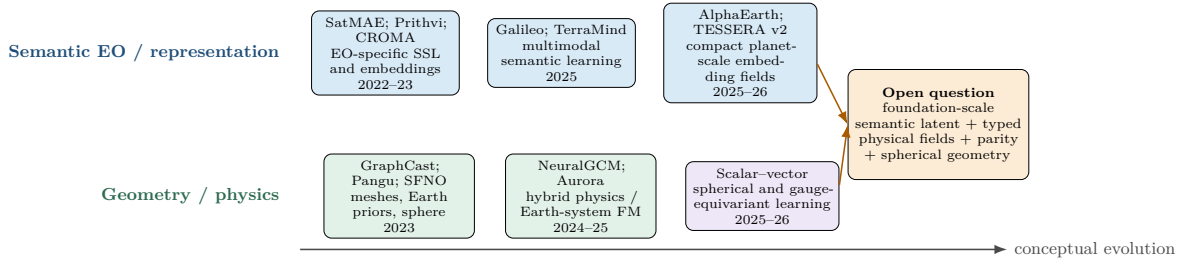
\begin{figure}[H]
\centering
\resizebox{0.98\textwidth}{!}{%
\begin{tikzpicture}[>=Latex,every node/.style={font=\scriptsize}]
\node[anchor=east,font=\small\bfseries\color{blueA}] at (0,2.85) {Semantic EO / representation};
\node[anchor=east,font=\small\bfseries\color{greenA}] at (0,0.25) {Geometry / physics};
\draw[->,thick,grayA] (0.25,-0.72)--(13.1,-0.72) node[right,font=\small]{conceptual evolution};

\node[draw,rounded corners,fill=blueB,align=center,text width=2.45cm,minimum height=1.25cm] at (1.8,2.85) {SatMAE; Prithvi; CROMA\\EO-specific SSL and embeddings\\2022--23};
\node[draw,rounded corners,fill=blueB,align=center,text width=2.45cm,minimum height=1.25cm] at (5.0,2.85) {Galileo; TerraMind\\multimodal semantic learning\\2025};
\node[draw,rounded corners,fill=blueB,align=center,text width=2.55cm,minimum height=1.25cm] at (8.25,2.85) {AlphaEarth; TESSERA v2\\compact planet-scale embedding fields\\2025--26};

\node[draw,rounded corners,fill=greenB,align=center,text width=2.5cm,minimum height=1.25cm] at (2.05,0.25) {GraphCast; Pangu; SFNO\\meshes, Earth priors, sphere\\2023};
\node[draw,rounded corners,fill=greenB,align=center,text width=2.5cm,minimum height=1.25cm] at (5.35,0.25) {NeuralGCM; Aurora\\hybrid physics / Earth-system FM\\2024--25};
\node[draw,rounded corners,fill=purpleB,align=center,text width=2.55cm,minimum height=1.25cm] at (8.65,0.25) {Scalar--vector spherical and gauge-equivariant learning\\2025--26};

\node[draw,rounded corners,fill=orangeB,align=center,text width=3.05cm,minimum height=2.0cm] (gap) at (11.85,1.55) {\textbf{Open question}\\foundation-scale semantic latent $+$ typed physical fields $+$ parity $+$ spherical geometry};
\draw[->,thick,orangeA] (9.65,2.65)--(gap.west);
\draw[->,thick,orangeA] (10.05,0.45)--(gap.west);
\end{tikzpicture}%
}
\caption{Literature-to-gap evolution. The proposed work sits at the intersection of two successful streams rather than replacing either: semantic EO foundation models and geometry/physics-aware Earth-system models.}
\label{fig:evolution}
\end{figure}

\section{Precisely stated research gap}

\begin{gapbox}{Defensible gap statement}
\textbf{Verified literature}: EO foundation models excel at transferable semantic embeddings; several Earth-system models exploit sphere-aware operators, graph meshes, Earth-specific priors, or hybrid physical solvers; specialized equivariant networks can process scalar and vector fields on spheres.\\[2pt]
\textbf{Inference from the surveyed architectures}: there is not yet a well-established, general-purpose Earth foundation model whose pretrained latent representation systematically preserves heterogeneous physical field types---including scalar, tangent polar-vector, parity-sensitive axial/pseudoscalar, and higher-order tensor channels---while jointly learning semantic geoembeddings across modalities.\\[2pt]
\textbf{Unresolved scientific question}: whether such typing provides measurable OOD generalization, data efficiency, physical consistency, or robustness beyond parameter-matched conventional models with correct geometric augmentation.
\end{gapbox}

This gap is important only for tasks in which transformation behavior is causally relevant. Land-cover classification from a north-up image may gain little. Wind/current forecasting, cyclone/vorticity diagnostics, transport, displacement estimation, ocean eddies, SAR polarimetric reasoning, and cross-projection transfer are more plausible beneficiaries.

\section{Research question, hypotheses, and objectives}

\subsection{Primary research question}
\begin{keybox}{Primary RQ}
\textbf{Does explicitly preserving the transformation laws of heterogeneous Earth-system fields within a transferable learned representation improve cross-region, cross-frame, low-data, and dynamical generalization compared with strong conventional foundation-model representations that treat physical components as ordinary channels?}
\end{keybox}

\subsection{Testable subquestions}
\begin{enumerate}
  \item \textbf{RQ1 -- necessity:} Are gains concentrated in tasks that genuinely depend on direction, rotation, circulation, anisotropy, or tensor structure?
  \item \textbf{RQ2 -- augmentation versus architecture:} Does exact/approximate equivariance outperform a conventional model trained with correctly transformed rotation/reflection augmentation at matched parameter count and compute?
  \item \textbf{RQ3 -- minimal intervention:} Can a lightweight typed equivariant adapter/input-output head recover most of the benefit of a fully equivariant backbone?
  \item \textbf{RQ4 -- decomposition:} Does Hodge/Helmholtz decomposition of vector fields add predictive or data-efficiency gains beyond direct typed-vector processing, and under what flow regimes?
\end{enumerate}

\subsection{Falsifiable hypotheses}
\begin{table}[H]
\centering
\caption{Hypotheses are stated so the project can fail cleanly. Numerical thresholds are proposed preregistration targets rather than literature-established constants.}
\label{tab:hypotheses}
\small
\begin{tabularx}{\textwidth}{p{1.0cm} X p{4.1cm}}
\toprule
& \textbf{Hypothesis} & \textbf{Falsification signal} \\
\midrule
$H_1$ & On held-out orientations and geographically disjoint regions, an $\OO(2)$-typed model will improve vector-sensitive forecast error by at least \textbf{5\% relative} to a parameter-matched, correctly augmented baseline, without materially degrading scalar targets. & $<5\%$ reproducible gain, confidence interval crossing zero, or scalar degradation $>2\%$. \\
$H_2$ & The advantage of typed equivariance will increase as training data are reduced and as coordinate-frame shift increases. & No interaction between model type and data fraction/frame perturbation. \\
$H_3$ & A typed adapter/head around a conventional backbone will recover at least \textbf{80\% of the OOD gain} of a fully equivariant model while requiring substantially less engineering/compute. & Adapter gains are small or require nearly the same complexity as full equivariance. \\
$H_4$ & Hodge/Helmholtz features will improve tasks tied to rotational/divergent flow structure, but will not consistently improve unrelated scalar tasks. & Uniform gains everywhere (suggesting generic extra-capacity confound) or no gains even on circulation-sensitive targets. \\
$H_0$ & After fair augmentation, parameter matching, and tuning, explicit field typing provides no practically or statistically meaningful improvement. & This is the null to be retained if $H_1$--$H_4$ fail. \\
\bottomrule
\end{tabularx}
\end{table}

\subsection{Research objectives}
\textbf{Overall objective:} determine when physically typed geometric representations are necessary in an Earth foundation model, and identify the minimum architecture that captures any demonstrated benefit.

\begin{enumerate}
  \item Formalize Earth variables as scalar, tangent-vector, pseudoscalar/axial, covector, and tensor fields under local coordinate transformations.
  \item Build controlled planar/tangent-frame experiments that isolate field-type equivariance from spherical-domain geometry.
  \item Quantify gains in OOD accuracy, equivariance defect, data efficiency, physical diagnostic fidelity, robustness, and compute cost.
  \item Compare explicit typing against strong augmentation, spherical operators, and Hodge/Helmholtz decomposition.
  \item If and only if the core hypothesis survives, integrate a lightweight typed module into a pretrained Earth model and evaluate cross-task transfer.
  \item Establish failure boundaries: tasks and regimes where generic embeddings are sufficient and geometric typing is unnecessary.
\end{enumerate}

\section{Candidate research formulations and ranking}

\begin{table}[H]
\centering
\caption{Ranking of candidate formulations. Scores are reasoned proposal judgments (1=low, 5=high), not empirical results. Score order is \textbf{N/V/F/R}: novelty / scientific value / feasibility / risk of gratuitous complexity.}
\label{tab:rank}
\small
\begin{tabularx}{\textwidth}{P{1.05cm} P{3.75cm} P{2.15cm} >{\raggedright\arraybackslash}X}
\toprule
\textbf{ID} & \textbf{Formulation} & \textbf{N/V/F/R} & \textbf{Assessment} \\
\midrule
A & Typed scalar/vector/axial/tensor tokens in a multimodal Transformer & 3 / 3 / 4 / 4 & Labels alone are insufficient: arbitrary attention/mixing can destroy vector or parity transformation laws. Useful only when fusion operations also respect type. \\
B & Gauge-/rotation-equivariant latent representations on the sphere & 5 / 5 / 2 / 3 & Highest long-term scientific value because it handles global geometry and local frames, but too expensive and technically coupled for the first falsification experiment. \\
C & Hodge/Helmholtz decomposition before tokenization & 3 / 3 / 4 / 2 & Cheap, interpretable, and physically meaningful for flow; best treated as an ablation because its benefit is likely task- and regime-dependent. \\
D & Equivariance-aware adapters/heads around an existing backbone & 4 / 4 / 5 / 1 & \textbf{Best first intervention}: low compute, clean causal attribution, compatible with existing FMs, and capable of falsifying the core hypothesis. \\
Hybrid & D $\rightarrow$ C ablation $\rightarrow$ B only after positive evidence & 5 / 5 / 4 / 2 & \textbf{Recommended research program}: increases mathematical complexity only after simpler experiments demonstrate real OOD or data-efficiency gains. \\
\bottomrule
\end{tabularx}
\end{table}

\begin{cautionbox}{Why formulation A alone is not enough}
A Transformer can be told that two channels are ``wind-$u$'' and ``wind-$v$,'' but if arbitrary linear layers independently mix those components, the representation does not necessarily transform as a vector. The research contribution must be in the \emph{transformation-consistent operations}, not only metadata labels.
\end{cautionbox}

\section{Proposed conceptual architecture}

The long-term architecture should be hybrid: generic invariant/semantic channels coexist with typed equivariant channels. In a local tangent plane, an $\OO(2)$ representation is attractive because reflections distinguish polar vectors from parity-sensitive pseudoscalars. For global modeling, gauge-equivariant or sphere-native representations are needed because a single globally consistent east/north basis cannot be treated as a flat Cartesian frame without distortion.

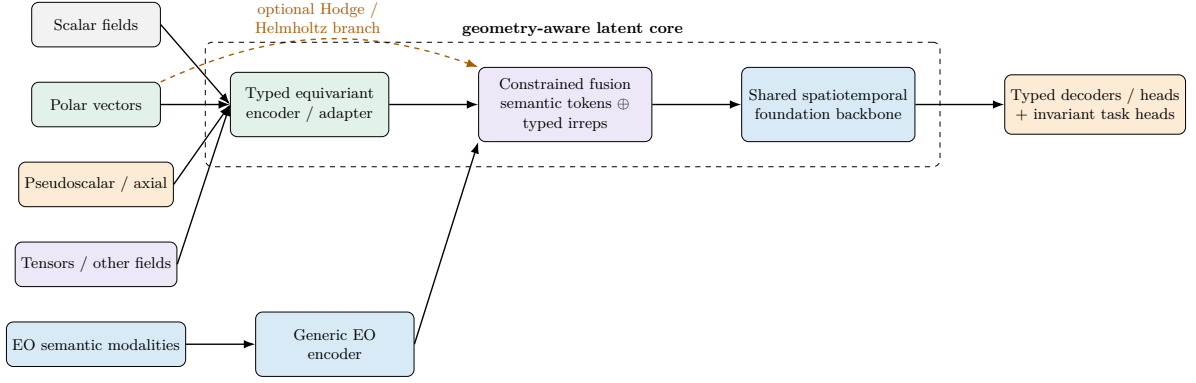
\begin{figure}[H]
\centering
\resizebox{0.98\textwidth}{!}{%
\begin{tikzpicture}[node distance=7mm and 7mm,>=Latex,every node/.style={font=\small}]
\node[draw,rounded corners,fill=grayB,minimum width=2.6cm,minimum height=0.9cm] (s) {Scalar fields};
\node[draw,rounded corners,fill=greenB,below=of s,minimum width=2.6cm,minimum height=0.9cm] (v) {Polar vectors};
\node[draw,rounded corners,fill=orangeB,below=of v,minimum width=2.6cm,minimum height=0.9cm] (p) {Pseudoscalar / axial};
\node[draw,rounded corners,fill=purpleB,below=of p,minimum width=2.6cm,minimum height=0.9cm] (t) {Tensors / other fields};
\node[draw,rounded corners,fill=blueB,below=of t,minimum width=2.6cm,minimum height=0.9cm] (eo) {EO semantic modalities};

\node[draw,rounded corners,fill=greenB,right=14mm of v,minimum width=3.2cm,minimum height=1.3cm,align=center] (typedenc) {Typed equivariant\\encoder / adapter};
\node[draw,rounded corners,fill=blueB,right=14mm of eo,minimum width=3.2cm,minimum height=1.3cm,align=center] (semenc) {Generic EO\\encoder};
\node[draw,rounded corners,fill=purpleB,right=18mm of typedenc,minimum width=3.5cm,minimum height=1.5cm,align=center] (fusion) {Constrained fusion\\semantic tokens $\oplus$\\typed irreps};
\node[draw,rounded corners,fill=blueB,right=18mm of fusion,minimum width=3.5cm,minimum height=1.5cm,align=center] (backbone) {Shared spatiotemporal\\foundation backbone};
\node[draw,rounded corners,fill=orangeB,right=18mm of backbone,minimum width=3.4cm,minimum height=1.2cm,align=center] (heads) {Typed decoders / heads\\+ invariant task heads};

\foreach \a in {s,v,p,t} {\draw[->,thick] (\a.east) -- (typedenc.west);}
\draw[->,thick] (eo.east) -- (semenc.west);
\draw[->,thick] (typedenc.east) -- (fusion.west);
\draw[->,thick] (semenc.east) -- (fusion.south west);
\draw[->,thick] (fusion.east) -- (backbone.west);
\draw[->,thick] (backbone.east) -- (heads.west);
\draw[->,dashed,thick,orangeA,bend left=22] (v.north east) to node[above,align=center]{optional Hodge /\\Helmholtz branch} (fusion.north west);
\node[draw,dashed,rounded corners,fit=(typedenc)(fusion)(backbone),inner sep=5mm,label={[font=\small\bfseries]above:geometry-aware latent core}] {};
\end{tikzpicture}}
\caption{Long-term hybrid architecture. The MVP should implement only the green ``typed adapter'' and typed output head around a small conventional backbone; full spherical/gauge-aware fusion is a later stage.}
\label{fig:architecture}
\end{figure}

\subsection{Transformation constraints}
For a local orthogonal frame transform $R\in\OO(2)$:
\begin{align}
  s' &= s,\\
  \vect v' &= R\vect v,\\
  \zeta' &= \det(R)\,\zeta,
\end{align}
where $s$ is scalar, $\vect v=(u,v)^\top$ is a tangent polar vector, and $\zeta$ is vertical vorticity treated as a pseudoscalar on the 2D surface. A model $f$ is equivariant when
\begin{equation}
  f(T_g x)=T_g f(x), \qquad g\in G,
\end{equation}
for the chosen transformation group and representation on each field type.

\section{End-to-end feasibility dry run: smallest experiment that can falsify the idea}

\subsection{Why ERA5 first}
ERA5/WeatherBench-style data are ideal for the first test because they contain both scalars and true directional fields, support large temporal sample sizes, and are standard in AI weather evaluation \citep{weatherbench2}. Starting directly from Sentinel imagery is scientifically weaker for this hypothesis because most optical bands are scalar radiometric measurements; the strongest case for geometric typing is a vector-sensitive dynamical task.

\subsection{Dry-run data design}
\begin{table}[H]
\centering
\caption{Proposed minimum variable set. Variables can be expanded only after the core test is stable.}
\label{tab:variables}
\small
\begin{tabularx}{\textwidth}{p{2.2cm} p{2.2cm} p{2.5cm} p{2.6cm} X}
\toprule
\textbf{Variable} & \textbf{Role} & \textbf{Geometric type} & \textbf{Transform under local $R\in\OO(2)$} & \textbf{Use} \\
\midrule
2 m temperature & input/output & scalar & invariant & scalar control target \\
mean sea-level pressure & input/output & scalar & invariant & scalar dynamics / cyclone context \\
10 m eastward + northward wind $(u_{10},v_{10})$ & input/output & tangent polar vector & $\vect v' = R\vect v$ & primary vector target \\
850 hPa wind $(u_{850},v_{850})$ & optional input/output & tangent polar vector & same & tests vertical consistency \\
500 hPa geopotential & optional input/output & scalar & invariant & large-scale circulation context \\
vertical relative vorticity $\zeta$ & derived diagnostic / optional target & 2D pseudoscalar (normal axial component) & $\zeta'=\det(R)\zeta$ & parity/circulation-sensitive test \\
horizontal divergence $\delta$ & derived diagnostic & scalar under planar rotations/reflections & invariant under orthogonal frame change & Hodge/flow diagnostic \\
\bottomrule
\end{tabularx}
\end{table}

\textbf{Spatial domain.} Use four climatically distinct, non-overlapping regional patch sets (for example North Atlantic/Europe, North America, Indian monsoon region, and Southern Hemisphere mid-latitudes). Reproject each patch into a local tangent/equal-area coordinate frame before applying arbitrary rotations/reflections. This avoids pretending that raw latitude--longitude pixels form a globally flat Euclidean plane.

\textbf{Temporal split.} A concrete initial protocol is 2010--2019 train, 2020 validation, 2021--2022 in-domain test, with additional 2023--2024 or later years reserved for temporal-shift testing where data access permits. Spatial OOD regions must be withheld entirely from training. The exact years may be adjusted to data availability, but the principle of temporal and geographic separation is non-negotiable.

\textbf{Resolution.} Begin at 1.5$^\circ$ or 1.0$^\circ$ for fast global/regional prototyping using WeatherBench-compatible preprocessing; move to 0.25$^\circ$ only after architecture correctness is established. This keeps the first experiment feasible on a single modern GPU.

\textbf{Forecast task.} Predict $t+6$ h from two preceding states $(t-6,t)$, then evaluate 6, 12, 24, and 48 h autoregressive rollouts. A second variant predicts increments rather than absolute values to reduce persistence dominance.

\subsection{Coordinate-correct augmentation}
For every rotation/reflection augmentation, the image grid and physical components must be transformed consistently. Rotating the raster without rotating $(u,v)$ is an invalid baseline. The augmented conventional model is deliberately strong because the scientific question is whether architectural equivariance offers value \emph{beyond} correct augmentation.

\subsection{Model ladder}
\begin{table}[H]
\centering
\caption{Minimal model ladder for causal attribution.}
\label{tab:models}
\small
\begin{tabularx}{\textwidth}{p{1.1cm} p{3.0cm} X p{3.0cm}}
\toprule
\textbf{ID} & \textbf{Model} & \textbf{Definition} & \textbf{Purpose} \\
\midrule
B0 & Conventional U-Net / compact ViT & All variables are ordinary scalar channels; no rotation/reflection augmentation. & Establish naive baseline. \\
B1 & Augmentation-matched baseline & Same architecture/parameters as B0; training uses correct $\OO(2)$ transforms of both grids and vector/pseudoscalar components. & Tests whether augmentation already solves the problem. \\
G1 & Typed adapter + typed head & Conventional backbone bracketed by $\OO(2)$-equivariant field-aware input/output modules; scalar/vector/pseudoscalar representations are explicit. & \textbf{Primary MVP}. Smallest intervention likely to falsify $H_1$. \\
G2 & Fully equivariant compact network & Group-equivariant blocks throughout, parameter matched as closely as possible. & Upper bound on benefit from full equivariance. \\
H1 & G1 + Hodge/Helmholtz features & Add rotational/divergent vector decomposition as extra physically meaningful channels. & Tests whether decomposition gives independent benefit. \\
S1 & Sphere-native comparison & A small SFNO or other spherical model on a coarse global grid. & Separates domain-geometry benefit from field-type benefit. \\
\bottomrule
\end{tabularx}
\end{table}

\subsection{Hodge/Helmholtz branch}
On a suitable planar domain, a smooth vector field can be decomposed conceptually as
\begin{equation}
\vect v = \nabla \phi + \nabla^{\perp}\psi + \vect h,
\end{equation}
where $\nabla\phi$ is curl-free, $\nabla^{\perp}\psi$ is divergence-free, and $\vect h$ is harmonic (boundary/topology dependent). In atmospheric flow, neither divergence nor rotation should be forced to zero; the decomposition is used as a representation, not an invalid physical constraint. Prior ML work shows that Helmholtz/Hodge structure can be embedded into learning pipelines for flow estimation \citep{hdnet}, supporting feasibility but not guaranteeing usefulness for weather.

\subsection{Losses}
Use a weighted multi-task objective:
\begin{equation}
\mathcal{L}=\lambda_v\mathcal{L}_{\text{vec}}+\lambda_s\mathcal{L}_{\text{scalar}}+\lambda_\zeta\mathcal{L}_{\text{vort}}+\lambda_{eq}\mathcal{L}_{\text{eq}}+\lambda_{spec}\mathcal{L}_{\text{spectral}},
\end{equation}
with the key rule that losses must be identical across comparable baselines except where a loss is itself the ablated intervention. A simple vector loss is
\begin{equation}
\mathcal{L}_{\text{vec}}=\E\left[(\hat u-u)^2+(\hat v-v)^2\right].
\end{equation}
An optional numerical equivariance regularizer for models that are not exactly equivariant is
\begin{equation}
\mathcal{L}_{\text{eq}}=\E_{g\sim G}\frac{\|f(T_gx)-T_gf(x)\|_2^2}{\|f(x)\|_2^2+\epsilon}.
\end{equation}

\subsection{Evaluation metrics and stress tests}
\begin{table}[H]
\centering
\caption{Evaluation must distinguish ordinary forecast skill from geometric correctness.}
\label{tab:metrics}
\small
\begin{tabularx}{\textwidth}{p{3.0cm} p{3.2cm} X}
\toprule
\textbf{Metric / test} & \textbf{Definition or example} & \textbf{What it diagnoses} \\
\midrule
Vector RMSE & {\footnotesize$\sqrt{\E[(\hat u-u)^2+(\hat v-v)^2]}$} & Main directional forecast accuracy. \\
Speed error & {\footnotesize$|\|\hat{\vect v}\|-\|\vect v\||$} & Separates speed from direction quality. \\
Angular error & circular angle between predicted and true wind & Direct directional fidelity; handle low-speed cases with thresholding. \\
Scalar RMSE / ACC & standard weather metrics & Checks that geometric machinery does not harm ordinary scalar prediction. \\
Vorticity / divergence error & compare derivatives/derived fields & Tests whether flow structure improves even when component RMSE changes little. \\
Equivariance defect & {\scriptsize$\|f(T_gx)-T_gf(x)\|/(\|f(x)\|+\epsilon)$} & Directly measures the property being claimed. \\
Kinetic-energy spectrum & spectral energy versus wavenumber & Detects oversmoothing or unphysical scale distortion. \\
Cross-region OOD & test on never-seen geographic regions & Tests transfer beyond memorized local climatology. \\
Frame stress test & random rotations/reflections with correct component transforms & Isolates coordinate dependence. \\
Low-data curves & train on 10/25/50/100\% data & Tests data efficiency predicted by symmetry priors. \\
Autoregressive rollout & 6--48 h initially; later longer & Detects whether local gains survive compounding. \\
\bottomrule
\end{tabularx}
\end{table}

\subsection{Statistical protocol}
Use paired evaluation on identical forecast initializations. Report bootstrap confidence intervals with resampling units chosen above the autocorrelation scale (e.g., blocks of dates) and stratified by held-out region. For the central claim, test the difference G1$-$B1 rather than G1$-$B0. Report both statistical significance and effect size. Hyperparameter search budgets should be matched, and the geometry-aware model should not receive more tuning trials simply because it is novel.

\subsection{Go/no-go thresholds}
\begin{successbox}{Proceed to foundation-model scale only if all core conditions are met}
\begin{enumerate}
  \item G1 improves at least two predeclared OOD vector metrics by $\ge 5\%$ relative to B1 across multiple held-out regions, with confidence intervals excluding zero.
  \item Equivariance defect is materially reduced (target: $\ge 50\%$ relative reduction) and the gain is not confined to synthetic rotations alone.
  \item Scalar-target degradation is $<2\%$ and no major stability failure appears in short autoregressive rollouts.
  \item Training/inference overhead for G1 is modest (target: $<25$--$35\%$ in the MVP), or the accuracy gain is large enough to justify more.
  \item Low-data and/or cross-region experiments show a consistent advantage; otherwise the symmetry prior may be solving only an artificial frame-perturbation test.
\end{enumerate}
\end{successbox}

\begin{cautionbox}{Stop or narrow the project if}
B1 closes essentially all of the gap; improvements occur only on transformed copies but not real geographic OOD data; geometry-aware modules improve metrics only by increasing parameter count; Hodge features help unrelated tasks equally (suggesting a capacity confound); or the computational overhead dominates practical benefit. A negative result is scientifically valuable because it establishes that standard embeddings plus augmentation are sufficient for the tested Earth regime.
\end{cautionbox}

\begin{figure}[H]
\centering
\resizebox{0.99\textwidth}{!}{%
\begin{tikzpicture}[node distance=6mm and 8mm,>=Latex,every node/.style={font=\small}]
\node[draw,rounded corners,fill=grayB,align=center,minimum width=2.4cm] (era) {ERA5 /\\WeatherBench2};
\node[draw,rounded corners,fill=grayB,right=of era,align=center,minimum width=2.7cm] (prep) {local tangent\\reprojection};
\node[draw,rounded corners,fill=greenB,right=of prep,align=center,minimum width=2.8cm] (types) {assign field types\\scalar / vector / parity};
\node[draw,rounded corners,fill=blueB,right=of types,align=center,minimum width=3.0cm] (train) {B0, B1, G1, G2, H1\\matched training};
\node[draw,rounded corners,fill=purpleB,right=of train,align=center,minimum width=2.7cm] (stress) {OOD + frame +\\low-data stress tests};
\node[draw,rounded corners,fill=orangeB,right=of stress,align=center,minimum width=2.5cm] (decision) {go / no-go\\decision};
\draw[->,thick] (era)--(prep);
\draw[->,thick] (prep)--(types);
\draw[->,thick] (types)--(train);
\draw[->,thick] (train)--(stress);
\draw[->,thick] (stress)--(decision);
\node[below=5mm of train,draw,dashed,rounded corners,fill=orangeB,align=center] (hodge) {optional Hodge/Helmholtz ablation};
\draw[->,dashed] (types.south) |- (hodge.west);
\draw[->,dashed] (hodge.east) -| (train.south);
\end{tikzpicture}%
}
\caption{End-to-end dry run. Complexity is introduced only after the core field-typing comparison is valid.}
\label{fig:dryrun}
\end{figure}
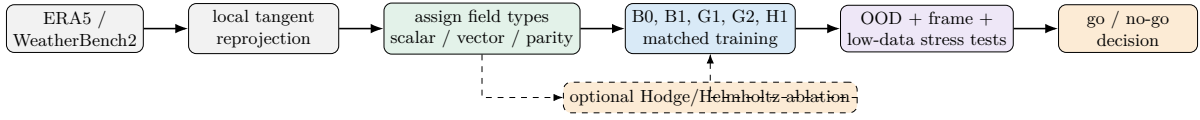

\section{Stage-2 bridge to an actual foundation model}

If the MVP succeeds, the next experiment should avoid training a billion-parameter model from scratch. Instead, freeze most of an existing pretrained Earth backbone and add a small geometry-aware interface. Two routes are plausible:

\begin{enumerate}
  \item \textbf{Earth-system route:} adapt an open weather/Earth-system backbone (e.g., Aurora-style encoder/decoder or a smaller spherical operator) so vector and parity-sensitive variables pass through typed adapters before/after the generic latent core. This directly tests dynamical transfer.
  \item \textbf{EO multimodal route:} fuse a pretrained EO semantic representation (e.g., AlphaEarth- or TESSERA-like embeddings, or an open model such as Galileo, TerraMind, or OlmoEarth) with typed ERA5, ocean, or current fields. The semantic embedding remains generic; the physical branch carries transformation-aware features.
\end{enumerate}

The second route has greater ``foundation model'' novelty because it combines mapping/perception semantics with physical dynamics, but it should come only after the first route shows the typed representation has value.

\section{Foundation-scale pretraining concept if the evidence is positive}

\subsection{Pretraining data mixture}
A future model could combine:
\begin{itemize}
  \item scalar EO fields: optical/radar-derived scalar channels, DEM, land surface temperature, precipitation;
  \item tangent vectors: wind, currents, displacement/velocity fields;
  \item parity-sensitive fields: vertical vorticity or other signed circulation quantities;
  \item tensors where scientifically justified: SAR covariance/coherency products, diffusion/strain/structure tensors;
  \item semantic modalities: text/geotagged descriptions, labels/pseudo-labels, generic image embeddings.
\end{itemize}

\subsection{Pretraining objectives}
A credible foundation objective should mix semantic and geometric goals:
\begin{align}
\mathcal{L}_{FM} ={}& \lambda_{mask}\mathcal{L}_{masked} + \lambda_{forecast}\mathcal{L}_{forecast}
+ \lambda_{cross}\mathcal{L}_{crossmodal} \\
& + \lambda_{eq}\mathcal{L}_{equivariance}
+ \lambda_{phys}\mathcal{L}_{physical\ diagnostics}.
\end{align}
The physical term should focus on diagnostics that are true identities or evaluation quantities, not impose false constraints. For example, atmospheric horizontal flow is not incompressible, so a generic divergence-free penalty would be scientifically wrong.

\subsection{Typed latent design}
One mathematically clean formulation is a direct sum of irreducible representation channels,
\begin{equation}
\mathcal{Z}=\mathcal{Z}_{\mathrm{sem}}\oplus \mathcal{Z}_{0}\oplus\mathcal{Z}_{1}\oplus\mathcal{Z}_{\mathrm{parity}}\oplus\mathcal{Z}_{2}\oplus\cdots,
\end{equation}
where $\mathcal{Z}_{\mathrm{sem}}$ contains generic semantic embeddings and $\mathcal{Z}_\ell$ contains fields transforming under chosen group representations. Tensor-product or gauge-equivariant interactions allow information exchange while preserving transformation rules \citep{tfn,gauge,nequip}.

\section{Ablation matrix}

\begin{table}[H]
\centering
\caption{Ablations required to attribute any improvement to geometry rather than extra parameters or data.}
\label{tab:ablations}
\small
\begin{tabularx}{\textwidth}{p{3.2cm} X p{4.3cm}}
\toprule
\textbf{Ablation} & \textbf{Comparison} & \textbf{Interpretation} \\
\midrule
No augmentation & B0 vs B1 & Quantifies how much correct augmentation alone helps. \\
Typed adapter only & G1 vs B1 & Primary causal test of explicit field typing. \\
Full equivariance & G2 vs G1 & Determines whether equivariance throughout the backbone is necessary. \\
No parity channel & remove pseudoscalar/axial representation & Tests whether reflection-sensitive structure adds value beyond rotation equivariance. \\
No vector typing & encode $(u,v)$ as two scalar channels within G architecture & Isolates benefit of the vector irrep itself. \\
Hodge decomposition & H1 vs G1 & Tests decomposition independently of equivariance. \\
Capacity match & widen B1 to match G1/G2 parameter count and FLOPs & Rules out trivial capacity explanation. \\
Loss match & remove auxiliary equivariance/diagnostic loss where not intrinsic & Separates architecture from regularization. \\
Synthetic-frame vs real OOD & rotated test sets vs withheld geographic regions & Prevents success only on artificial transformations. \\
Scalar-only task & predict temperature/pressure without vector target & Establishes boundary where typing should not help much. \\
\bottomrule
\end{tabularx}
\end{table}

\section{Risk register and mitigation}

\begin{table}[H]
\centering
\caption{Main scientific and implementation risks.}
\label{tab:risks}
\small
\begin{tabularx}{\textwidth}{p{3.5cm} p{2.2cm} X}
\toprule
\textbf{Risk} & \textbf{Severity} & \textbf{Mitigation} \\
\midrule
Coordinate-frame mistakes create fake gains & Very high & Unit-test every transformation with analytic vector fields; verify inverse transforms and equivariance numerically before training. \\
Reflection semantics are mishandled & High & Treat 2D vorticity as pseudoscalar; distinguish $\SO(2)$ from $\OO(2)$; document parity conventions. \\
Atmospheric flow is forced into inappropriate constraints & High & Use Hodge components as features/diagnostics, not hard divergence-free assumptions. \\
Equivariant model has more parameters/compute & High & Parameter/FLOP matching and widened conventional controls. \\
Spherical geometry confounds field typing & High & Begin in local tangent patches; add sphere-native model as separate factor later. \\
Gains vanish on real OOD data & High & Accept null; do not scale to FM. Publish boundary conditions if analysis is rigorous. \\
Foundation backbone is too expensive to modify & Medium & Use adapters, frozen encoders, low-resolution data, and smaller open backbones first. \\
EO tasks are mostly scalar/semantic & Medium & Target vector-sensitive applications first; do not claim universal improvement. \\
Higher-order tensors are included without a real use case & Medium & Add tensor fields only after identifying a task such as SAR polarimetry or stress/strain where tensor structure is native. \\
\bottomrule
\end{tabularx}
\end{table}

\section{Expected contributions}

If the hypothesis is supported, the work can make five nontrivial contributions:
\begin{enumerate}
  \item \textbf{Conceptual:} a clean taxonomy separating semantic embeddings from physically transforming geometric fields in GeoAI.
  \item \textbf{Methodological:} a minimal typed adapter architecture that can retrofit existing Earth models instead of requiring a new FM from scratch.
  \item \textbf{Empirical:} controlled evidence about when equivariance beats correct augmentation under cross-region and low-data shifts.
  \item \textbf{Physical:} explicit treatment of vector and parity-sensitive quantities, with Hodge/flow diagnostics and transformation stress tests.
  \item \textbf{Foundation-model direction:} a validated path toward hybrid semantic + typed Earth representations rather than speculative complexity.
\end{enumerate}

If the hypothesis is rejected, the research can still contribute a valuable negative result: for the tested Earth regimes, generic embeddings plus correct augmentation are sufficient, and full typed equivariance is not worth its compute/engineering cost.

\section{Staged roadmap}

\begin{table}[H]
\centering
\caption{Evidence-gated roadmap. Time is deliberately expressed as stages rather than promises of calendar duration.}
\label{tab:roadmap}
\small
\begin{tabularx}{\textwidth}{p{1.2cm} p{3.0cm} X p{3.2cm}}
\toprule
\textbf{Stage} & \textbf{Deliverable} & \textbf{Core work} & \textbf{Gate to next stage} \\
\midrule
0 & Mathematical validation & transformation/unit tests for scalars, vectors, pseudoscalars; synthetic advection/rotation data & numerical equivariance tests pass to tolerance \\
1 & ERA5 local dry run & B0/B1/G1, spatial/temporal OOD, low-data curves & $H_1$ shows practical gain vs B1 \\
2 & Mechanism ablations & G2, parity ablation, Hodge branch, spectrum/flow diagnostics & benefit is attributable and not capacity-only \\
3 & Sphere/global validation & HEALPix/gauge or sphere-native model; compare domain geometry vs field typing & gains persist globally and across latitude/frame changes \\
4 & Pretrained-backbone adapter & geometry-aware module on a frozen/open Earth backbone & transfer gain survives pretrained semantic latent \\
5 & Multimodal FM prototype & EO embeddings + weather/ocean typed fields; multi-task pretraining & clear cross-task benefits justify scale-up \\
6 & Foundation-scale study & larger pretraining, scaling laws, downstream benchmark suite & only if earlier evidence remains positive \\
\bottomrule
\end{tabularx}
\end{table}

\section{Recommended thesis/paper framing}

\begin{keybox}{Working title}
\textbf{Do Earth Foundation Models Need Physically Typed Representations? A Controlled Study of Scalar, Vector, Parity-Sensitive and Semantic Geospatial Features}
\end{keybox}

An alternative, more ambitious title after positive MVP evidence is:
\begin{quote}
\textbf{Geometry-Aware Earth Foundation Models: Joint Semantic and Equivariant Representation Learning for Heterogeneous Geophysical Fields}
\end{quote}

The first title is scientifically safer because it foregrounds the empirical question rather than assuming the answer.

\section{What would constitute a high-impact outcome?}
A high-impact result is not merely a few-percent gain on a rotated benchmark. The strongest outcome would be a consistent pattern: typed models need fewer data, maintain correct behavior under unseen coordinate frames, improve real cross-region generalization, preserve vector/vorticity structure during rollout, and can be attached to an existing pretrained Earth model with modest overhead. That would establish a new design principle for Earth foundation models. Conversely, showing convincingly that augmentation-matched generic embeddings erase the advantage would also be valuable because it prevents a large amount of unnecessary architectural complexity.

\section{Conclusion}
The literature supports a serious but carefully bounded research opportunity. EO foundation models such as AlphaEarth, TESSERA, Prithvi, Clay, Galileo, TerraMind, and OlmoEarth demonstrate the strength of semantic embeddings \citep{alphaearth,tessera2,prithvi2,clay,galileo,terramind,olmoearth}. Weather and climate models such as GraphCast, Pangu-Weather, SFNO, Aurora, and NeuralGCM show that Earth geometry and physics can matter profoundly \citep{graphcast,pangu,sfno,aurora,neuralgcm}. Geometric deep learning proves that scalar, vector, parity, and tensor features can be propagated with exact or principled transformation laws \citep{tfn,gauge,nequip}, while recent scalar/vector spherical networks show that this direction is technically feasible \citep{ballerin}. The unanswered question is whether these ideas remain useful at the level of general-purpose, multimodal Earth representation learning. The proposed program answers that question with escalating evidence rather than escalating complexity: \textbf{adapter first, full geometry later, foundation-scale pretraining only after the core hypothesis survives.}

\end{document}